\documentclass[10pt,twocolumn,letterpaper]{article}

\usepackage{iccv}
\usepackage{times}
\usepackage{epsfig}
\usepackage{graphicx}
\usepackage{amsmath}
\usepackage{amssymb}

\usepackage{booktabs} 

\usepackage{relsize} 
\usepackage{bbm} 

\usepackage[pagebackref=true,breaklinks=true,letterpaper=true,colorlinks,bookmarks=false]{hyperref}

\iccvfinalcopy 
\def\iccvPaperID{2122} 

\ificcvfinal\fi
\begin{document}

\title{RoomNet: End-to-End Room Layout Estimation}

\author{Chen-Yu Lee \quad Vijay Badrinarayanan \quad Tomasz Malisiewicz \quad Andrew Rabinovich
\vspace{1mm}
\\Magic Leap, Inc.\\
\tt\small{ \{clee,\hspace{0.3mm}vbadrinarayanan,\hspace{0.3mm}tmalisiewicz,\hspace{0.3mm}arabinovich\}@magicleap.com }
}

\maketitle

\begin{abstract}
This paper focuses on the task of room layout estimation from a monocular RGB image. Prior works break the problem into two sub-tasks: semantic segmentation of floor, walls, ceiling to produce layout hypotheses, followed by an iterative optimization step to rank these hypotheses. 

In contrast, we adopt a more direct formulation of this problem as one of estimating an ordered set of room layout keypoints. The room layout and the corresponding segmentation is completely specified given the locations of these ordered keypoints. We predict the locations of the room layout keypoints using RoomNet, an end-to-end trainable encoder-decoder network. On the challenging benchmark datasets Hedau and LSUN, we achieve state-of-the-art performance along with 200$\times$ to 600$\times$ speedup compared to the most recent work. Additionally, we present optional extensions to the RoomNet architecture such as including recurrent computations and memory units to refine the keypoint locations under the same parametric capacity. 
\end{abstract}

\section{Introduction}
\vspace{-1mm}
Room layout estimation from a monocular image, which aims to delineate a 2D boxy representation of an indoor scene, is an essential step for a wide variety of computer vision tasks, and has recently received great attention from several applications. These include indoor navigation~\cite{mirowski2016learning}, scene reconstruction/rendering~\cite{izadinia2016im2cad}, and augmented reality~\cite{xiao2014reconstructing, liu2015rent3d, detone2016deep}.

\begin{figure}[t]
\begin{center}
\includegraphics[width=0.95\linewidth]{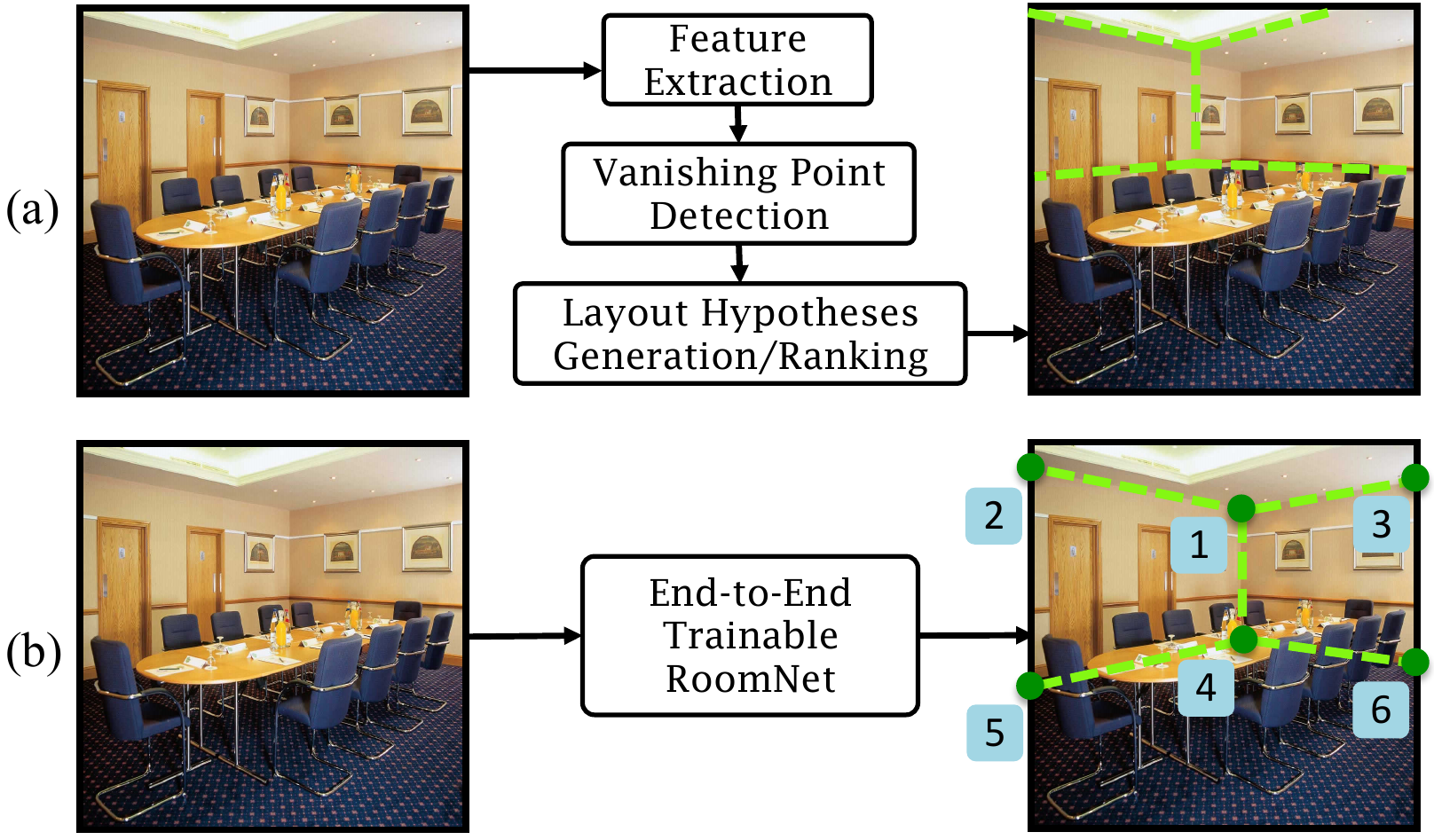}
\end{center}
    \vspace{-3mm}
   \caption{(a) Typical multi-step pipeline for room layout estimation. (b) Room layout estimation with RoomNet is direct and simple: run RoomNet,  extract a set of room layout keypoints, and connect the keypoints in a specific order to obtain the layout.}
\label{fig:pipeline}
\vspace{-1mm}
\end{figure}

The field of room layout estimation has been primarily focused on using bottom-up image features such as local color, texture, and edge cues followed by vanishing point detection.
A separate post-processing stage is used to clean up feature outliers and generate/rank a large set of room layout hypotheses with structured SVMs or conditional random fields (CRFs)~\cite{hedau2009recovering, gupta2010estimating, hedau2012recovering, ramalingam2013manhattan, zhang2013estimating}. 
In principle, the 3D reconstruction of the room layout can be obtained (up to scale) with knowledge of the 2D layout and the vanishing points. However, in practice, the accuracy of the final layout prediction often largely depends on the quality of the extracted low-level image features, which in itself is susceptible to local noise, scene clutter and occlusion.

\begin{figure*}[ht]
\begin{center}
\includegraphics[width=0.55\linewidth]{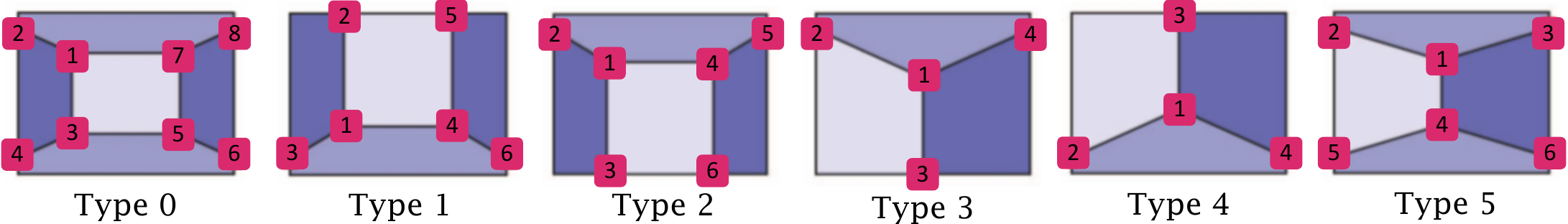}
\includegraphics[width=0.43\linewidth]{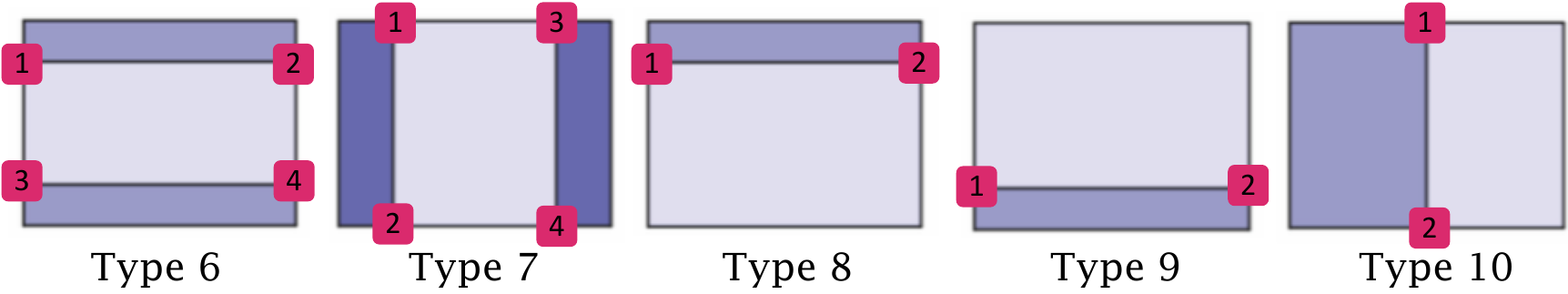}
\end{center}
    \vspace{-4mm}
   \caption{Definition of room layout types. The type is indexed from 0 to 10 as in~\cite{zhang2016large}. The number on each keypoint defines the specific order of points saved in the ground truth. For a given room type, the ordering of keypoints specifies their connectivities.}
\label{fig:layout_type}
\vspace{-0mm}
\end{figure*}

Recently, with the rapid advances in deep convolutional neural networks (CNNs) for semantic segmentation ~\cite{chen2014semantic, long2015fully, noh2015learning, badrinarayanan2015segnet}, researchers have been exploring the possibility of using such CNNs for room layout estimation.
More specifically, Mallya \etal~\cite{mallya2015learning} first train a fully convolutional network (FCN)~\cite{long2015fully} model to produce ``informative edge maps'' that replace hand engineered low-level image feature extraction. The predicted edge maps are then used to sample vanishing lines for layout hypotheses generation and ranking. 
Dasgupta \etal~\cite{dasgupta2016delay} use the FCN to learn semantic surface labels such as left wall, front wall, right wall, ceiling, and ground. Then connected components and hole filling techniques are used to refine the raw per pixel prediction of the FCN, followed by the classic vanishing point/line sampling methods to produce room layouts. 
However, despite the improved results, these methods use CNNs to generate a new set of ``low-level'' features and fall short of exploiting the end-to-end learning ability of CNNs. In other words, the raw CNN predictions need to be post-processed by an expensive hypotheses testing stage to produce the final layout. This, for example, takes the pipeline of Dasgupta \etal~\cite{dasgupta2016delay} 30 seconds to process each frame.

In this work, we address the problem top-down by directly training CNNs to infer both the room layout corners (keypoints) and room type. Once the room type is inferred and the corresponding set of ordered keypoints are localized, we can connect them in a specific order to obtain the 2D spatial room layout. 
The proposed method, RoomNet, is direct and simple as illustrated in Figure~\ref{fig:pipeline}: The network takes an input image of size 320 $\times$ 320, processes it through a convolutional encoder-decoder architecture, extracts a set of room layout keypoints, and then simply connects the obtained keypoints in a specific order to draw a room layout. 
The semantic segmentation of the layout surfaces is simply obtainable as a consequence of this connectivity.

Overall, we make several contributions in this paper: (1) reformulate the task of room layout estimation as a keypoint localization problem that can be directly addressed using CNNs, (2) a custom designed convolutional encoder-decoder network, RoomNet, for parametrically efficient and effective joint keypoint regression and room layout type classification, and (3) state-of-the-art performance on challenging benchmarks Hedau~\cite{hedau2009recovering} and LSUN~\cite{zhang2016large} along with 200$\times$ to 600$\times$ speedup compared to the most recent work.

\begin{figure*}[t]
\begin{center}
\includegraphics[width=0.9\linewidth]{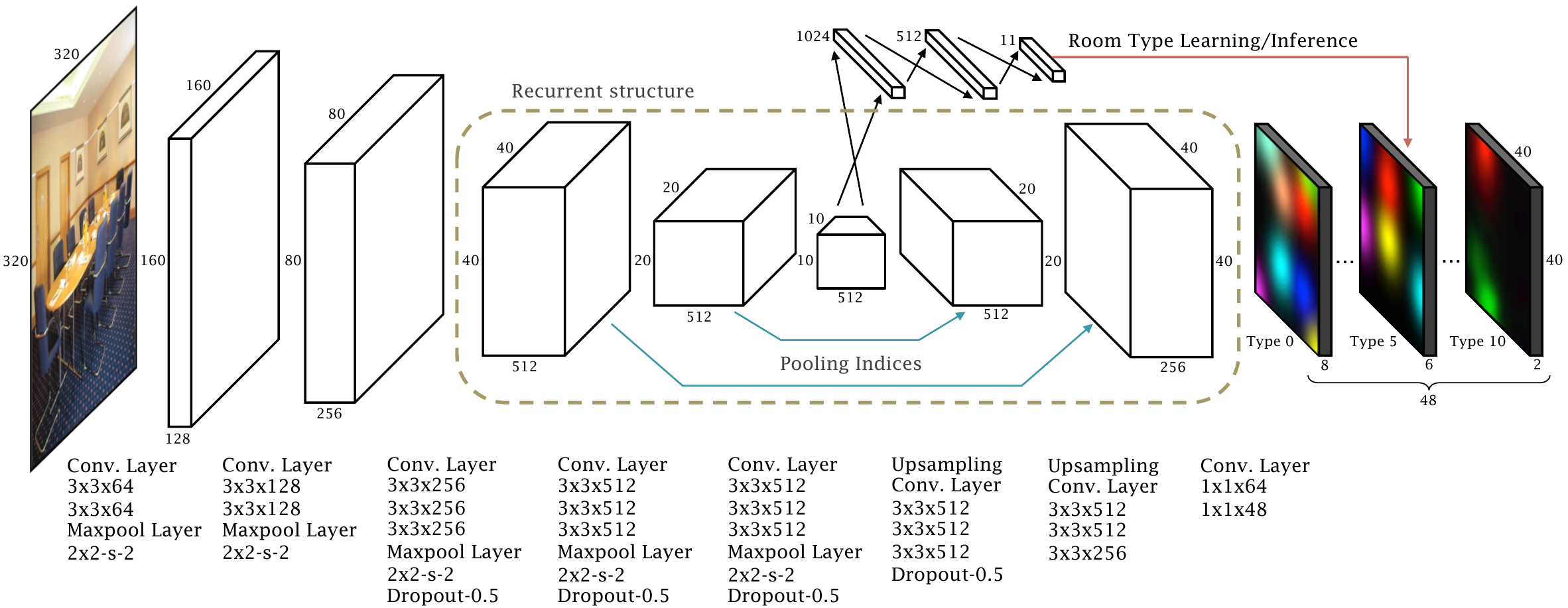}
\end{center}
    \vspace{-2mm}
   \caption{An illustration of the RoomNet base architecture. A decoder upsamples its input using the transferred pooling indices from its encoder to produce sparse feature maps followed by a several convolutional layers with trainable filter banks to densify the feature responses. The final decoder output keypoint heatmaps are fed to a regressor with Euclidean losses. A side head with 3 fully-connected layers is attached to the bottleneck layer and used to train/predict the room type class label, which is then used to select the associated set of keypoint heatmaps. The full model of RoomNet with recurrent encoder-decoder (center dashed line block) further performs keypoint refinement as shown in Figure~\ref{fig:unrolled} (b) and~\ref{fig:refinement}.}
\label{fig:pipeline_full}
\vspace{-2mm}
\end{figure*}

\section{RoomNet}
\vspace{-1mm}
\subsection{Keypoint-based room layout representaiton}
To design an effective room layout estimation system, it is important to choose a proper target output representation
that is end-to-end trainable and can be inferred efficiently.
Intuitively, one can assign geometric context classes (floor, walls, and ceiling) to each pixel, and then try to obtain room layout keypoints and boundaries based on the pixel-wised labels. However, it is non-trivial to derive layout keypoints and boundaries from the raw pixel output.
In contrast, if we can design a model that directly outputs a set of ordered room layout keypoint locations,
it is then trivial to obtain both keypoint-based and pixel-based room layout representations.

Another important property of using a keypoint-based representation is that it eliminates the ambiguity in the pixel-based representation. Researchers have shown that CNNs often have difficulty distinguishing between different surface identities. For instance, CNNs can be confused between the front wall class and the right wall class, and thereby output irregular or mixed pixel-wise labels within the same surface -- this is well illustrated by Figure 5 and 6 from~\cite{dasgupta2016delay}. This phenomenon also largely undermines the overall room layout estimation performance.

Hence, we propose to use a keypoint-based room layout representation to train our model. Figure~\ref{fig:layout_type} shows a list of room types with their respective keypoint definition as defined by~\cite{zhang2016large}. These 11 room layouts cover most of the possible situations under typical camera poses and common cuboid representations under ``Manhattan world assumption''~\cite{yuille2000manhattan}. Once the trained model predicts correct keypoint locations with an associated room type, we can then simply connect these points in a specific order to produce boxy room layout representation.

\subsection{Architecture of RoomNet}	\label{variants}
We design a CNN to delineate room layout structure using 2D keypoints. 
The input to the network is a single RGB image and the output of the network is a set of 2D keypoints in a specific order with an associated room type.

\vspace{1.5mm}
\noindent \textbf{Keypoint estimation}
The base network architecture for keypoint estimation is inspired by the recent successes in the field of semantic segmentation~\cite{long2015fully, noh2015learning, badrinarayanan2015segnet}. Here we adopt the SegNet architecture proposed by Badrinarayanan \etal~\cite{badrinarayanan2015segnetarxiv, badrinarayanan2015segnet} with modifications. 
Initially designed for segmentation, the SegNet framework consists of encoder and decoder sub-networks -- the encoder of the SegNet maps an input image to lower resolution feature maps, and then the role of the decoder is to upsample the low resolution encoded feature maps to full input resolution for pixel-wise classification. In particular, the decoder uses pooling indices computed in the max-pooling step of the corresponding encoder to perform non-linear upsampling. This eliminates the need for learning to upsample. The upsampled maps are sparse and are convolved with trainable filters to produce dense feature map.
This architecture has proven to provide good performance with competitive inference time and efficient memory usage as compared to other recent semantic segmentation architectures.

The base architecture of RoomNet adopts essentially the same convolutional encoder-decoder network as in SegNet. It takes an image of an indoor scene and directly outputs a set of 2D room layout keypoints to recover the room layout structure. Each keypoint ground truth is represented by a 2D Gaussian heatmap centered at the true keypoint location as one of the channels in the output layer.\footnote{We color-code and visualize multiple keypoint heatmaps in a single 2D image in Figure~\ref{fig:pipeline_full}, Figure~\ref{fig:refinement} and the rest of the paper.} The encoder-decoder architecture processes the information flow through bottleneck layers, enforcing it to implicitly model the relationship among the keypoints that encode the 2D structure of the room layout.

The decoder of the RoomNet upsamples the feature maps from the bottleneck layer with spatial dimension 10 $\times$ 10 to 40 $\times$ 40 instead of the full resolution 320 $\times$ 320 as shown in Figure~\ref{fig:pipeline_full}. This is because we empirically found that using the proposed 2D keypoint-based representation can already model the room layout effectively at 40 $\times$ 40 scale  (results are similar as compared to training decoder sub-network at full resolution). Using this ``trimmed'' decoder sub-network also significantly reduces the memory/time cost during both training and testing due to the high computation cost of convolution at higher resolutions.

\vspace{1.5mm}
\noindent \textbf{Extending to multiple room types}
The aforementioned keypoint estimation framework serves as a basic room layout estimation system for one particular room type. To generalize this approach for multiple room types, one possible solution is to train one network  per class as in the Single Image 3D Interpreter Network of Wu \etal~\cite{wu2016single}. However, in order to maximize efficiency, we design the RoomNet to be fast from the ground up. Encouraged by the recent object detection works YOLO~\cite{redmon2016you} and SSD~\cite{liu2016ssd} that utilize a single neural network to predict bounding boxes and class probabilities directly from full images in one evaluation, our proposed RoomNet similarly predicts room layout keypoints and the associated room type with respect to the input image in one forward pass. To achieve this goal, we increase the number of channels in the output layer to match the total number of keypoints for all 11 room types (total 48 keypoints for 11 room types derived from Figure~\ref{fig:layout_type}), and we also add a side head with fully connected layers to the bottleneck layer (the layer where usually used for image classification) for room type prediction as shown in Figure~\ref{fig:pipeline_full}.

We denote a training example as $(\boldsymbol{\mathcal{I}}, \boldsymbol{y}, t)$, where $\boldsymbol{y}$ stands for the ground truth coordinates of the $k$ keypoints with room type $t$ for the input image $\boldsymbol{\mathcal{I}}$.
At training stage, we use the Euclidean loss as the cost function for layout keypoint heatmap regression and use the cross-entropy loss for the room type prediction.
Given the keypoint heatmap regressor $\varphi$ (output from the decoder sub-network), and the room type classifier $\psi$  (output from the fully-connected side head layer), we can then optimize the following loss function:
\begin{equation}
\footnotesize
\sum_{k}
     \mathlarger{\mathbbm{1}}_{k,t}^{\text{keypoint}}
            \|
		G_k(\boldsymbol{y}) - \varphi_k(\boldsymbol{\mathcal{I}})
            \|^2
	- \lambda \sum_{c} \mathlarger{\mathbbm{1}}_{c,t}^{\text{room}} \log(\psi_c(\boldsymbol{\mathcal{I}}))       
\end{equation}
where $\mathbbm{1}_{k,t}^{\text{keypoint}}$ denotes if keypoint $k$ appears in ground truth room type $t$, 
$\mathbbm{1}_{c,t}^{\text{room}}$ denotes if room type index $c$ equals to the ground truth room type $t$,
$G$ is a Gaussian centered at $\boldsymbol{y}$ and the weight term $\lambda$ is set to 5 by cross validation.
The first term in the loss function compares the predicted heatmaps to ground-truth heatmaps synthesized for each keypoint separately. The ground truth for each keypoint heatmap is a 2D Gaussian centered on the true keypoint location with standard deviation of 5 pixels as in the common practice in recent keypoint regression works~\cite{tompson2014joint, pfister2015flowing, carreira2016human, wu2016single}. The second term in the loss function encourages the side head fully-connected layers to produce a high confidence value with respect to the correct room type class label.

Note that one forward pass of the proposed architecture will produce keypoint heatmaps for all room types. However, the loss function only penalizes Euclidean regression error if the keypoint $k$ is present for the ground truth room type $t$ in the current input image $\boldsymbol{\mathcal{I}}$, effectively using the predicted room type indices to select the corresponding set of keypoint heatmaps to update the regressor. The same strategy applies at the test stage \ie the predicted room type is used to select the corresponding set of keypoint heatmaps in the final output.

\vspace{1.5mm}
\noindent \textbf{RoomNet extension for keypoint refinement}
Recurrent neural networks (RNNs) and its variant Long Short-Term Memory (LSTM)~\cite{hochreiter1997long} have proven to be extremely effective models when dealing with sequential data. Since then, researchers have been exploring the use of recurrent structures for static input format as well, such as recurrent convolutional layers~\cite{liang2015recurrent} and convLSTM layers~\cite{xingjian2015convolutional}.

Recently, more sophisticated iterative/recurrent architectures have been proposed for 2D static input, such as FCN with CRF-RNN~\cite{zheng2015conditional}, iterative error feedback networks~\cite{carreira2016human}, recurrent CNNs~\cite{belagiannis2016recurrent}, stacked encoder-decoder~\cite{newell2016stacked}, and recurrent encoder-decoder networks~\cite{peng2016recurrent, lee2016recursive}. These evidence show that adopting the ``time series'' concept when modeling a static input can also significantly improve the ability of the network to integrate contextual information and to reduce prediction error.

\begin{figure}[t]
\begin{center}
\includegraphics[width=1\linewidth]{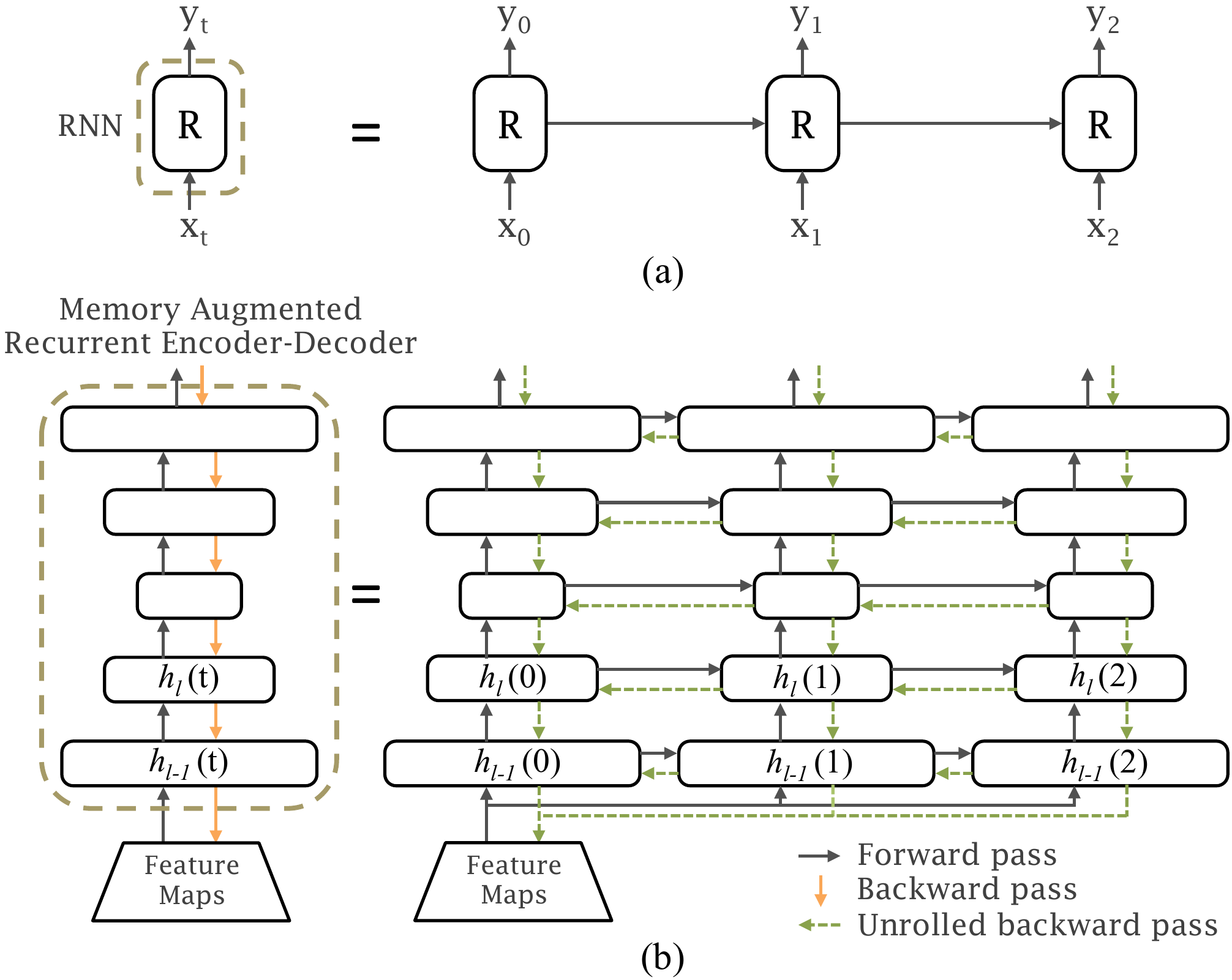}
\end{center}
    \vspace{-4mm}
   \caption{Illustration of unrolled (3 iterations) version of (a) a RNN and (b)
    the proposed memory augmented recurrent encoder-decoder architecture that mimics the behavior of a RNN but which is designed for a static input. Both structures have hidden units to store previous activations that help the inference at the current time step.}
\label{fig:unrolled}
\end{figure}

\begin{figure}[t]
\begin{center}
\includegraphics[width=0.98\linewidth]{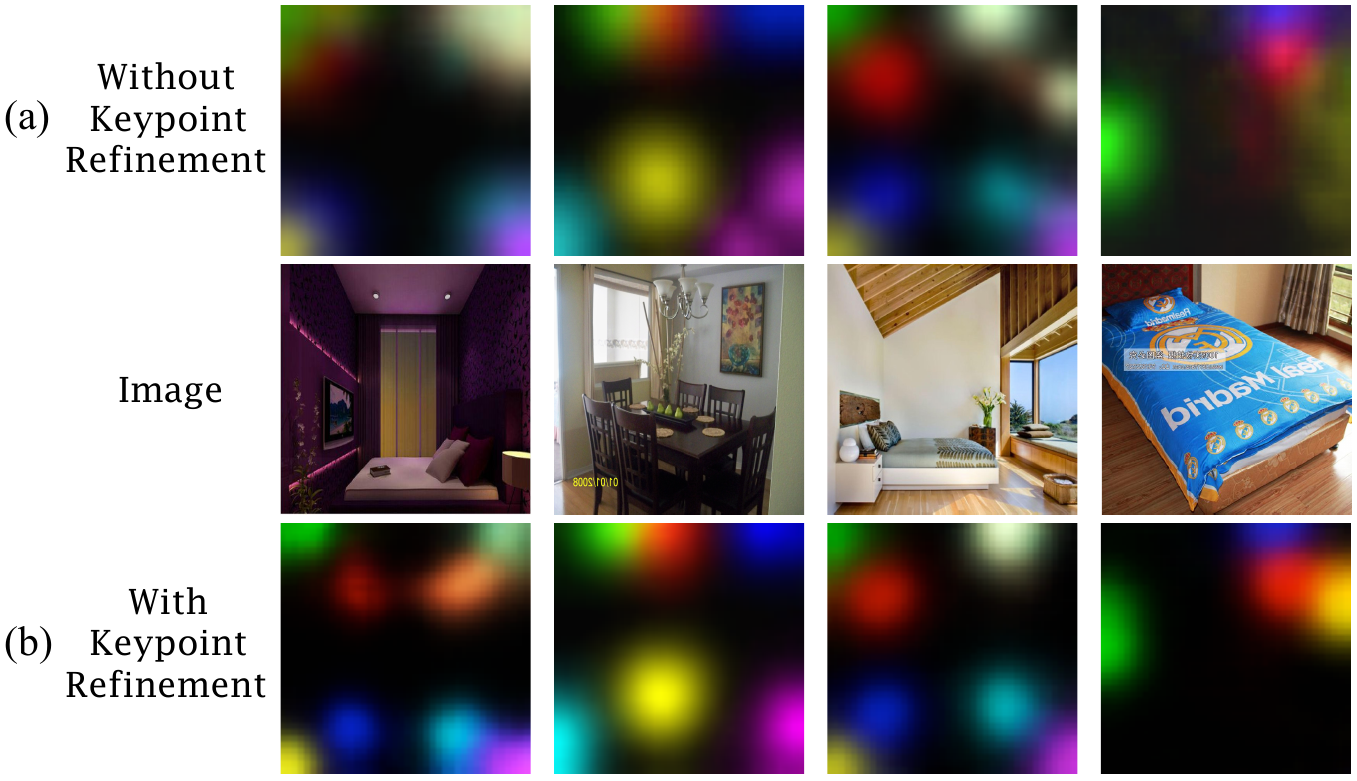}
\end{center}
    \vspace{-4mm}
   \caption{Room layout keypoint estimation from a single image (a) without refinement and (b) with refinement. Keypoint heatmaps from multiple channels are color-coded and shown in a single 2D image for visualization purposes. The keypoint refinement step produces more concentrated and cleaner heatmaps and removes some false positives.}
\label{fig:refinement}
\end{figure}

Motivated by the aforementioned successes, we extend our base RoomNet architecture by making the central encoder-decoder component (see center dashed line block in Figure~\ref{fig:pipeline_full}) recurrent. Specifically, we propose a memory augmented recurrent encoder-decoder (MRED) structure (see Figure~\ref{fig:unrolled} (b)) whose goal is to mimic the behavior of a typical recurrent neural network (Figure~\ref{fig:unrolled} (a)) in order to refine the predicted keypoint heatmaps over ``time'' -- the artificial time steps created by the recurrent structure.

Each layer in this MRED structure shares the same weight matrices through different time steps that convolve (denoted as $*$ symbol) with the incoming feature maps from the previous prediction $\mathbf{h}_{l}(t-1)$ at time step $t-1$ in the same layer $l$ and the current input $\mathbf{h}_{l-1}(t)$ at time step $t$ in the previous layer $l-1$, generating output at time step $t$ as:
\begin{equation}
\footnotesize
\begin{aligned}
\label{eq:2}
    \mathbf{h}_{l}(t)=
   \begin{cases}
    \sigma( \mathbf{w}_{l}^{\text{current}}  * \mathbf{h}_{l-1}(t) + \mathbf{b}_{l})  & \text{, $t=0$ } \\
    \sigma( \mathbf{w}_{l}^{\text{current}} * \mathbf{h}_{l-1}(t) + \mathbf{w}_{l}^{\text{previous}} * \mathbf{h}_{l}(t-1) + \mathbf{b}_{l})  & \text{, $t > 0$ } \\
   \end{cases} 
\end{aligned}
\end{equation}
where $\mathbf{w}_{l}^{\text{current}}$ and $\mathbf{w}_{l}^{\text{previous}}$ are the input and feed-forward weights for layer $l$. $\mathbf{b}_{l}$ is the bias for layer $l$. $\sigma$ is the ReLU activation function~\cite{nair2010rectified}.

Figure~\ref{fig:unrolled} (b) demonstrates the overall process of the information flow during forward- and backward- propagations through depth and time within the recurrent encoder-decoder structure. The advantages of using the proposed MRED architecture are: (1) exploiting the contextual and structural knowledge among keypoints iteratively through hidden/memory units (that have not been explored in recurrent convolutional encoder-decoder structure) and (2) weight sharing of the convolutional layers in the recurrent encoder-decoder, resulting in a much deeper network with a fixed number of parameters.
After refinement, the heatmaps of keypoints are much cleaner as shown in Figure~\ref{fig:refinement}. 
It is also interesting to observe the mistakes made early on and corrected later by the network (see third and fourth columns in Figure~\ref{fig:refinement}). We analyze the performance with and without the keypoint refinement step in Section~\ref{deep_supervision}, and we also evaluate different encoder-decoder variants in Section~\ref{discussion}.

\begin{figure}[t]
\begin{center}
\includegraphics[width=1\linewidth]{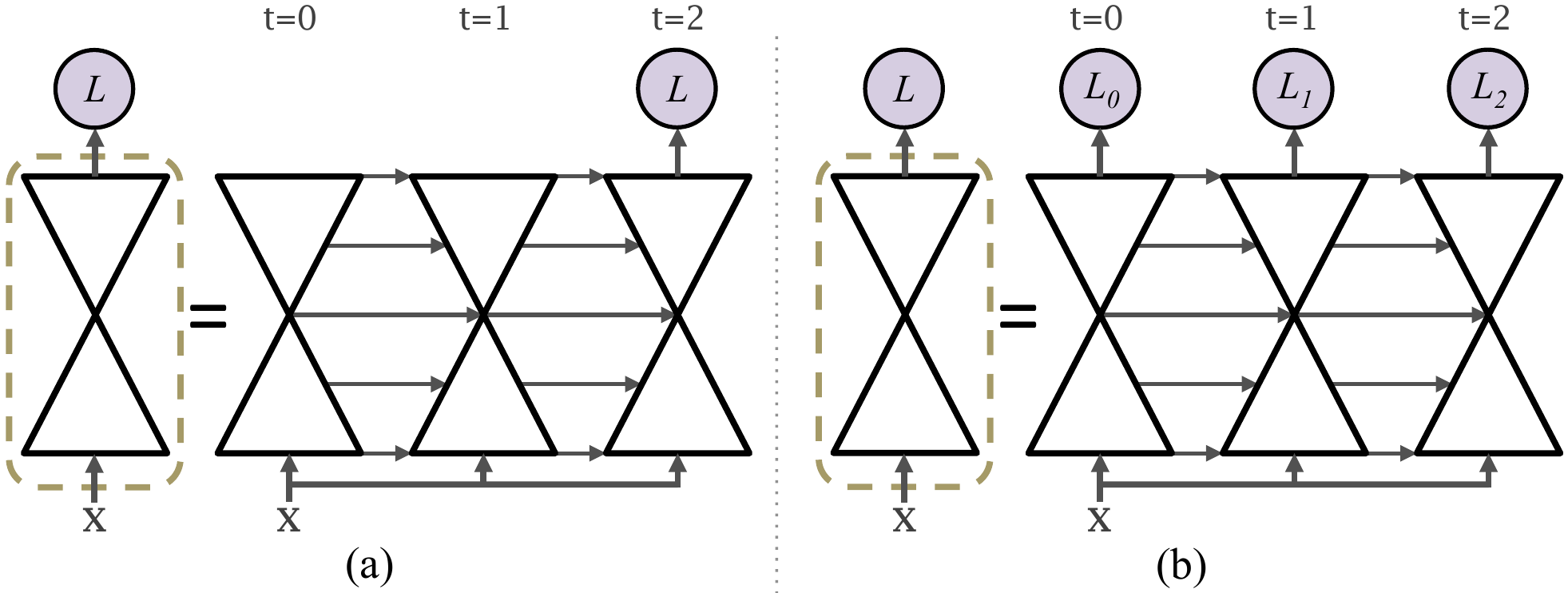}
\end{center}
    \vspace{-4mm}
   \caption{Illustration of the proposed memory augmented recurrent encoder-decoder architecture (a) without deep supervision through time and (b) with deep supervision through time.}
\label{fig:deep_super}
\vspace{-2mm}
\end{figure}

\vspace{1.5mm}
\noindent \textbf{Deep supervision through time}
When applying stacked, iterative, or recurrent convolutional structures, each layer in the network receives gradients across more layers or/and time steps, resulting in models that are much harder to train. For instance, the iterative error feedback network~\cite{carreira2016human} requires multi-stage training and the stacked encoder-decoder structure in~\cite{newell2016stacked} uses intermediate supervision at the end of each encoder-decoder even when batch normalization~\cite{ioffe2015batch} is used.
Following the practices in~\cite{lee2015deeply, szegedy2015going}, we extend the idea by injecting supervision at the end of each time step. The same loss function $L$ is applied to all the time steps as demonstrated in Figure~\ref{fig:deep_super}. Section~\ref{deep_supervision} and Table~\ref{tab:RoomNet_ablation2} provide details of the analysis and effect of the deep supervision through time.

\section{Experiments}
\vspace{-1mm}
\subsection{Datasets}
\vspace{-0.5mm}
We evaluate the proposed RoomNet framework on two challenging benchmark datasets: Hedau~\cite{hedau2009recovering} dataset and Large-scale Scene Understanding Challenge (LSUN) room layout dataset~\cite{zhang2016large}. The Hedau dataset contains 209 training, 53 validation, and 105 test images that are collected from the web and from LabelMe~\cite{russell2008labelme}. The LSUN dataset consists of 4000 training, 394 validation, and 1000 test images that are sampled from SUN database~\cite{xiao2010sun}. 
We follow the same experimental setup as Dasgupta \etal~\cite{dasgupta2016delay}. We rescale all input images to 320 $\times$ 320 pixels and train our network from scratch on the LSUN training set only. All experimental results are computed using the LSUN room layout challenge toolkit~\cite{zhang2016large} on the original image scales.

\subsection{Implementation details}
The input to the network is an RGB image of resolution 320 $\times$ 320 and the output is the room layout keypoint heatmaps of resolution 40 $\times$ 40 with an associated room type class label.
We apply the backpropagation through time (BPTT) algorithm to train the models with batch size 20 SGD, 0.5 dropout rate, 0.9 momentum, and 0.0005 weight decay. Initial learning rate is 0.00001 and decreased by a factor of 5 twice at epoch 150 and 200, respectively. All variants use the same scheme with 225 total epochs. The encoder and decoder weights are all initialized using the technique described in He \etal~\cite{he2015delving}. Batch normalization~\cite{ioffe2015batch} and ReLU~\cite{nair2010rectified} activation function are also used after each convolutional layer to improve the training process. We apply horizontal flipping of input images during training as the only data augmentation. The system is implemented in the open source deep learning framework Caffe~\cite{jia2014caffe}.

In addition, a ground truth keypoint heatmap has zero value (background) for most of its area and only a small portion of it corresponds to the Gaussian distribution (foreground associated with actual keypoint location). The output of the network therefore tends to converges to zero due to the imbalance between foreground and background distributions. For this reason, it is crucial to weight the gradients based on the ratio between foreground and background area for each keypoint heatmap. In our experiment, we degrade the gradients of background pixels by multiplying them with a factor of 0.2 and found this makes training significantly more stable.

Training from scratch takes about 40 hours on 4 NVIDIA Titan X GPUs. One forward inference of the full model (RoomNet recurrent 3-iter) takes 83 ms on a single GPU. For generating final test predictions we run both the original input and a flipped version of the image through the network and average the heatmaps together (accounting for a 0.12\% average improvement on keypoint error and a 0.15\% average improvement on pixel error) as in~\cite{newell2016stacked}. The keypoint location is chosen to be the max activating location of the corresponding heatmap.

\subsection{Results}
Two standard room layout estimation evaluation metrices are: (1) pixel error: pixelwise error between the predicted surface labels and ground truth labels, and (2) keypoint error: average Euclidean distance between the predicted keypoint and annotated keypoint locations, normalized by the image diagonal length.

\vspace{1.5mm}
\noindent \textbf{Accuracy}
We summarize the performance on both datasets in Table~\ref{tab:RoomNet_hedau} and~\ref{tab:RoomNet_lsun}. The previous best method is the two-step framework (per pixel CNN-based segmentation with a separate hypotheses ranking approach) Dasgupta \etal~\cite{dasgupta2016delay}. The proposed RoomNet significantly improves upon the previous results on both keypoint error and pixel error, achieving state-of-the-art performance \footnote{The side head room type classifier obtained 81.5\% accuracy on LSUN dataset.}.

To decouple the performance gains due to external data, we also prepare results of fine-tuning the RoomNet from a SUN~\cite{song2015sun} pre-trained model (on semantic segmentation task) and this achieves 6.09\% keypoint error and 9.04\% pixel error as compared of method in~\cite{ren2016coarse}\footnote{The multi-step method in~\cite{ren2016coarse} utilizes additional Hedau+~\cite{mallya2015learning} training set and fine-tunes from NYUDv2 RGBD~\cite{gupta2013perceptual} pre-trained models.} with 7.95\% keypoint error and 9.31\% pixel error on LSUN dataset.

\vspace{1.5mm}
\noindent \textbf{Runtime and complexity}
Efficiency evaluation on the input image size of 320 $\times$ 320 is shown in Table~\ref{tab:speed}. Our full model (RoomNet recurrent 3 iteration) achieves 200$\times$ speedup compares to the previous best method in~\cite{dasgupta2016delay}, and the base  RoomNet without recurrent structure (RoomNet basic) achieves 600$\times$ speedup. Note that the timing is for two forward passes as described earlier. Using either one of the proposed architecture can provide significant inference time reduction and an improved accuracy as shown in Table~\ref{tab:RoomNet_ablation1}.

\begin{table}
\footnotesize{
\setlength{\tabcolsep}{3pt}
\def\arraystretch{1.2}
\center
\begin{tabular}{@{}l c}
\toprule
\textbf{Method}  									&   Pixel Error (\%)  	\\
\midrule
Hedau \etal (2009)~\cite{hedau2009recovering}	      		&	21.20	\\ 
Del Pero \etal (2012)~\cite{del2012bayesian}				&	16.30	\\ 
Gupta \etal (2010)~\cite{gupta2010estimating}	      			&	16.20	\\ 
Zhao \etal (2013)~\cite{zhao2013scene}	      				&	14.50	\\ 
Ramalingam \etal (2013)~\cite{ramalingam2013manhattan}	&	13.34	\\ 
Mallya \etal (2015)~\cite{mallya2015learning} 	      			&	12.83	\\  
Schwing \etal (2012)~\cite{schwing2012efficient} 			&	12.8	\\  
Del Pero \etal (2013)~\cite{del2013understanding} 			&	12.7	\\  
Dasgupta \etal (2016)~\cite{dasgupta2016delay}     			&	9.73		\\
\midrule	
RoomNet	 recurrent 3-iter (ours)	      				     				& \textbf{8.36}	\\
\bottomrule
\end{tabular}
\vspace{0.2cm}
\caption{Performance on Hedau dataset~\cite{hedau2009recovering}. We outperform the previous best result in~\cite{dasgupta2016delay} using the proposed end-to-end trainable RoomNet.}
\label{tab:RoomNet_hedau}}
\end{table}

\begin{table}
\footnotesize{
\setlength{\tabcolsep}{3pt}
\def\arraystretch{1.2}
\center
\begin{tabular}{@{}l c c}
\toprule
\textbf{Method}  							& Keypoint Error (\%)  &   Pixel Error (\%)  	\\
\midrule
Hedau \etal (2009)~\cite{hedau2009recovering}	&  15.48     		&	24.23 \\ 
Mallya \etal (2015)~\cite{mallya2015learning} 	      	&  11.02      		&	16.71 \\  
Dasgupta \etal (2016)~\cite{dasgupta2016delay}     	&  8.20      		&	10.63 \\
\midrule	
RoomNet	 recurrent 3-iter (ours)	      						& \textbf{6.30}     	& \textbf{9.86}\\
\bottomrule
\end{tabular}
\vspace{0.2cm}
\caption{Performance on LSUN dataset~\cite{zhang2016large}. We outperform the previous best result in~\cite{dasgupta2016delay} on both keypoint and pixel errors using the proposed end-to-end trainable RoomNet.}
\label{tab:RoomNet_lsun}}
\end{table}

\begin{table}
\footnotesize{
\setlength{\tabcolsep}{3pt}
\def\arraystretch{1.2}
\center
\begin{tabular}{@{}l c c c}
\toprule
\textbf{Method}  				& FPS   	\\
\midrule
Del Pero \etal (2013)~\cite{del2013understanding}  & 0.001 \\
Dasgupta \etal (2016)~\cite{dasgupta2016delay} & 0.03 \\
RoomNet	 recurrent 3-iter  		& 5.96      		\\
RoomNet	 recurrent 2-iter		  	& 8.89       	\\
RoomNet	 basic  				& 19.26       	\\
\bottomrule
\end{tabular}
\vspace{0.2cm}
\caption{Runtime evaluation on an input size of 320$\times$320. The proposed RoomNet full model (3-iter) achieves 200$\times$ speedup and the basic RoomNet model achieves 600$\times$ speedup than the previous best method in~\cite{dasgupta2016delay}.}
\label{tab:speed}}
\vspace{-0.2cm}
\end{table}

\subsection{Analyzing RoomNet} \label{deep_supervision}
In this section, we empirically investigate the effect of each component in the proposed architecture with the LSUN dataset as our running example.

\begin{figure*}[ht]
\begin{center}
\includegraphics[width=0.75\linewidth]{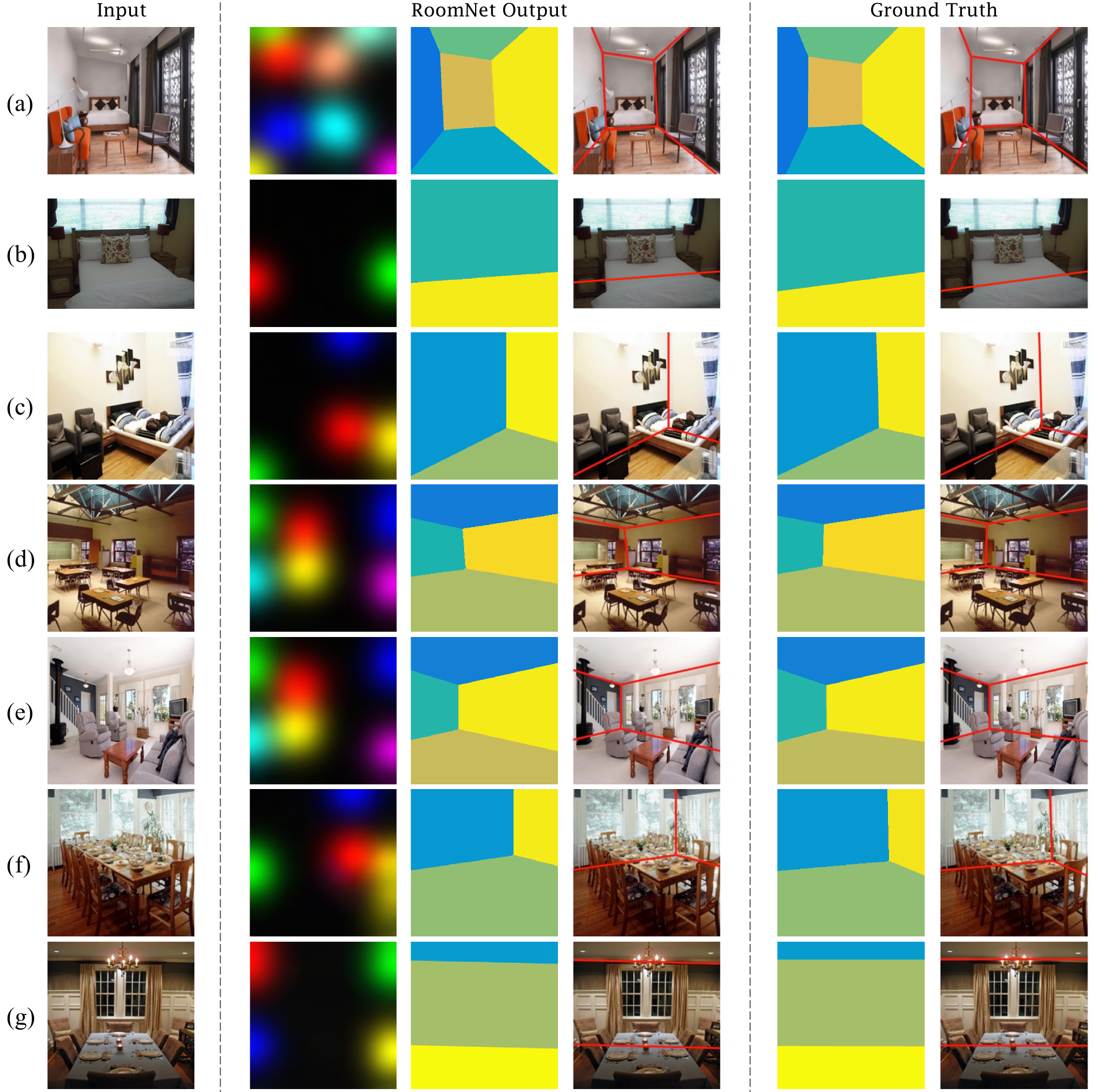}
\end{center}
\vspace{-3mm}
\caption{The RoomNet predictions and the corresponding ground truth on LSUN dataset. The proposed architecture takes a RGB input (first column) and produces room layout keypoint heatmaps (second column). The final keypoints are obtained by extracting the location with maximum response from the heatmaps. The third and fourth columns show a boxy room layout representation by simply connecting obtained keypoints in a specific order as in Figure~\ref{fig:layout_type}. The fifth and sixth columns show the ground truth.
Our algorithm is robust to keypoint occlusion by objects (ex: tables, chairs, beds).}
\label{fig:example1}
\vspace{-1mm}
\end{figure*}

\vspace{1.5mm}
\noindent \textbf{Recurrent vs direct prediction}
Table~\ref{tab:RoomNet_ablation1} shows the effectiveness of extending the RoomNet-basic architecture to a memory augmented recurrent encoder-decoder networks. We observed that more iterations led to lower error rates on both keypoint error and pixel error: the RoomNet with recurrent structure that iteratively regresses to correct keypoint locations achieves 6.3\% keypoint error and 9.86 pixel error as compared to the RoomNet without recurrent structure which achieves 6.95\% keypoint error and 10.46 pixel error. No further significant performance improvement is observed after 3 iterations. Notice that the improvement essentially came from the same parametric capacity within the networks since the weights of convolutional layers are shared across iterations.

\vspace{1.5mm}
\noindent \textbf{Importance of deep supervision through time}
When applying a recurrent structure with encoder-decoder architectures, each layer in the network receives gradients not only across depth but also through time steps between the input and the final objective function during training. It is therefore of interest to investigate the effect of adding auxiliary loss functions at different time steps. Table~\ref{tab:RoomNet_ablation2} demonstrates the impact of deep supervision through time using RoomNet with 2 and 3 recurrent iterations. We observed immediate reduction in both keypoint error and pixel error by adding auxiliary losses for both cases. This can be understood by the fact that the learning problem with deep supervision is much easier~\cite{lee2015deeply, szegedy2015going} through different time steps. It is also interesting to point out that RoomNet 3-iter performs worse than RoomNet 2-iter when deep supervision through time is not applied. This is rectified when deep supervision through time is applied. Overall, we validate that with more iterations in the recurrent structure, there is a stronger need to apply deep supervision through time to successfully train the proposed architecture.

\begin{figure*}[ht]
\begin{center}
\includegraphics[width=0.75\linewidth]{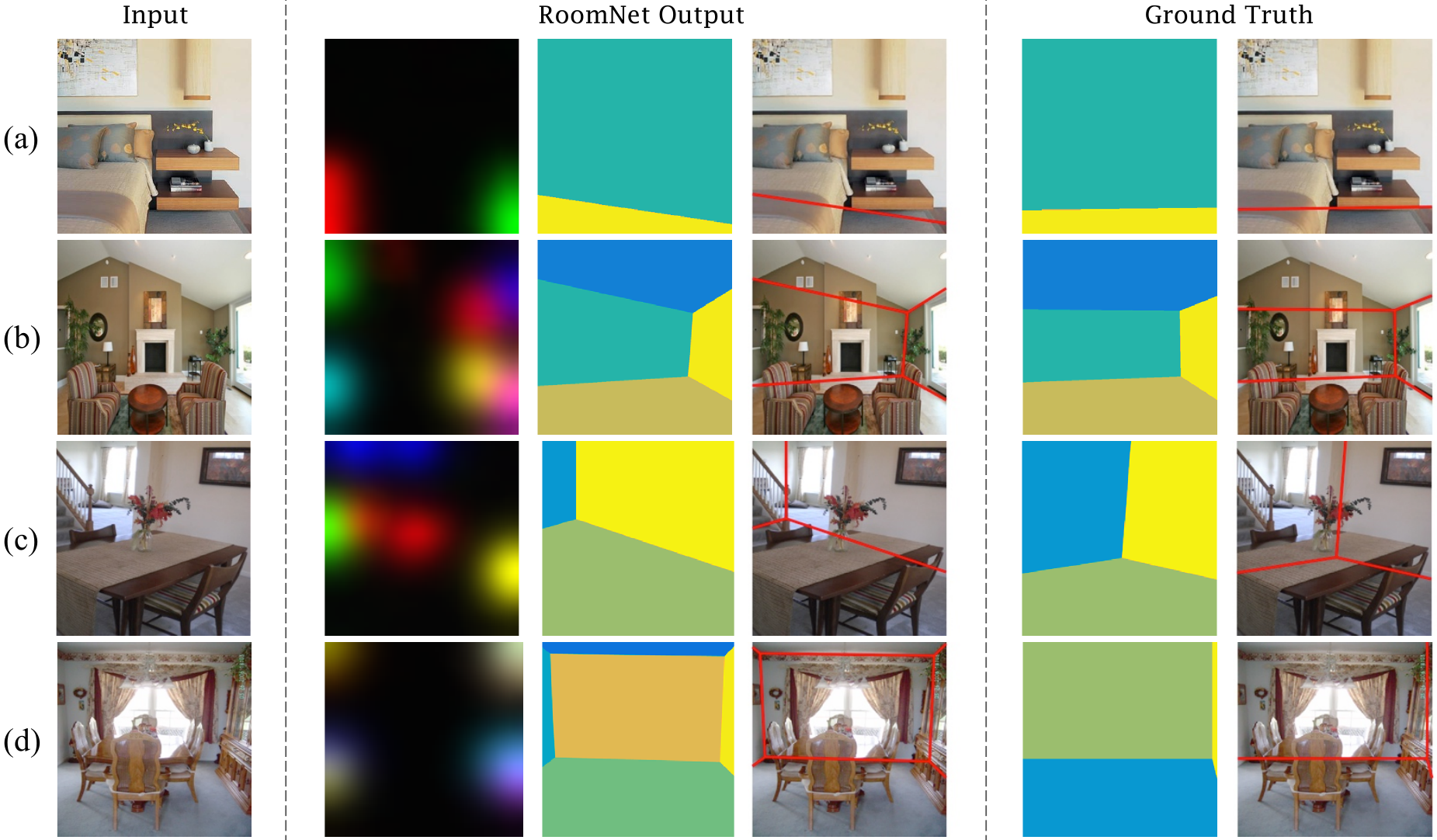}
\end{center}
    \vspace{-3mm}
   \caption{The ambigous cases where the RoomNet predictions do not match the human-annotated ground truth. The first column is the input image, the second column is predicted keypoint heatmaps, the third and fourth columns are obtained boxy representation, and the fifth and sixth columns show the ground truth.}
\label{fig:example2}
\vspace{-2mm}
\end{figure*}

\begin{table}
\footnotesize{
\setlength{\tabcolsep}{3pt}
\def\arraystretch{1.2}
\center
\begin{tabular}{@{}l c c}
\toprule
\textbf{Model}  					& Keypoint Error (\%)  	&   Pixel Error (\%)  	\\
\midrule
RoomNet	 basic				&  6.95    				&  10.46			\\ 
RoomNet	 recurrent 2-iter    		&  6.65      			&  9.97			\\	
RoomNet	 recurrent 3-iter	      		& \textbf{6.30}     		& \textbf{9.86}		\\
\bottomrule
\end{tabular}
\vspace{0.2cm}
\caption{The impact of keypoint refinement step (see Section~\ref{variants}) using the proposed memory augmented recurrent encoder-decoder architecture on LSUN dataset~\cite{zhang2016large}.}
\label{tab:RoomNet_ablation1}}
\end{table}

\begin{table}
\footnotesize{
\setlength{\tabcolsep}{3pt}
\def\arraystretch{1.2}
\center
\begin{tabular}{@{}l c c}
\toprule
\textbf{Model}  			& Keypoint Error (\%)	&   Pixel Error (\%)	\\
\midrule
RoomNet	 recurrent 2-iter  		&        				&   	\\
- w/o deep supervision through time    	&  6.93      			&  10.44	\\
- w/ deep supervision through time   	&  6.65      			&  9.97	\\
\midrule
RoomNet	 recurrent 3-iter  		&       				&   	\\
- w/o deep supervision through time   	&  6.95      			&  10.47	\\
- w/ deep supervision through time   	& \textbf{6.30}     		& \textbf{9.86}	\\
\bottomrule
\end{tabular}
\vspace{0cm}
\caption{The impact of deep supervision through time on LSUN dataset~\cite{zhang2016large} for RoomNets with 2 and 3 recurrent iterations.}
\label{tab:RoomNet_ablation2}}
\end{table}

\vspace{1.5mm}
\noindent \textbf{Qualitative results}
We show qualitative results of the proposed RoomNet in Figure~\ref{fig:example1}. When the image is clean and the room layout boundaries/corners are not occluded, our algorithm can recover the boxy room layout representation with high accuracy. Our framework is also robust to keypoint occlusion by objects (ex: tables, chairs, beds), demonstrated in Figure~\ref{fig:example1} (b)(c)(d)(f). The major failure cases are when room layout boundaries are barely visible (Figure~\ref{fig:example2} (a)(c)) or when there are more than one plausible room layout explanations for a given image of a scene (Figure~\ref{fig:example2} (b)(d)).

\section{Discussion}	\label{discussion}
\begin{figure}[ht]
\begin{center}
\includegraphics[width=1\linewidth]{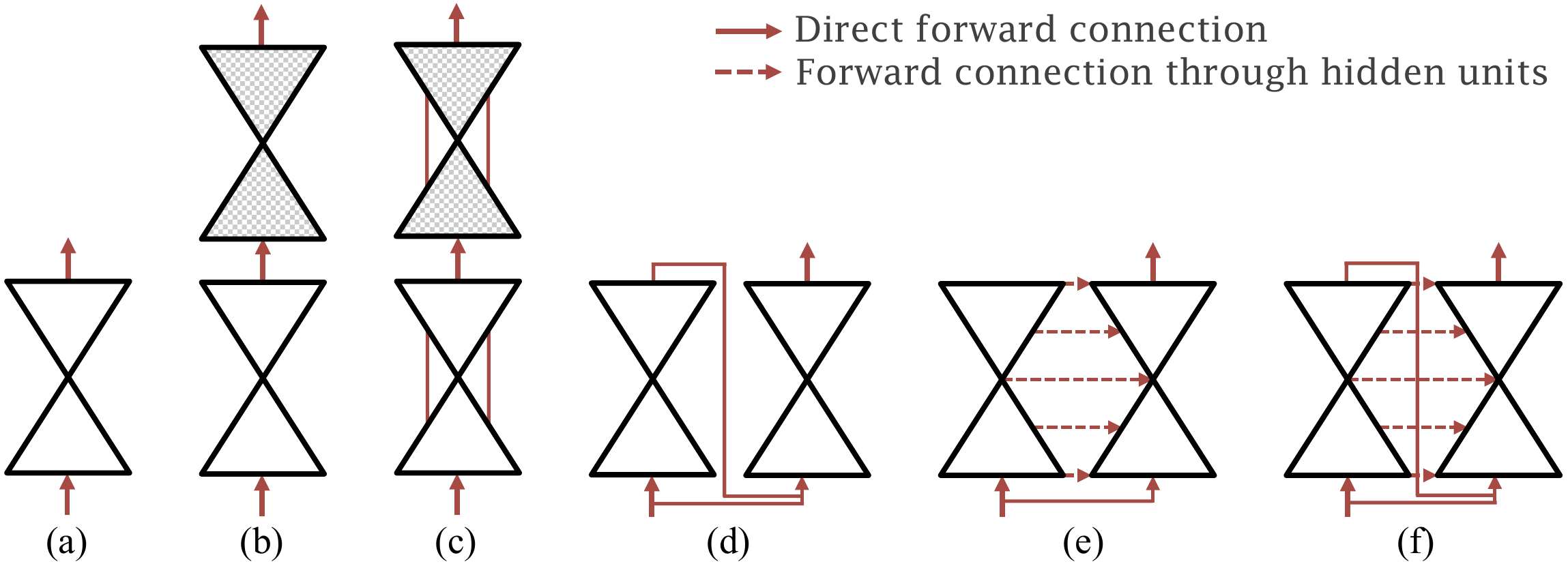}
\end{center}
    \vspace{-3mm}
   \caption{Illustration of different encoder-decoder architecture configurations: (a) vanilla encoder-decoder; (b) stacked encoder-decoder; (c) stacked encoder-decoder with skip-connections; (d) encoder-decoder with feedback; (e) memory augmented recurrent encoder-decoder; (f) memory augmented recurrent encoder-decoder with feedback.}
\label{fig:variants}
\vspace{-1mm}
\end{figure}

\vspace{-1mm}
\noindent \textbf{Alternative encoder-decoders}
We provide an evaluation of alternative encoder-decoder architectures for the room layout estimation task including: (a) a vanilla encoder-decoder (RoomNet basic), (b) stacked encoder-decoder, (c) stacked encoder-decoder with skip-connections; (d) encoder-decoder with feedback; (e) memory augmented recurrent encoder-decoder (RoomNet full); (f) memory augmented recurrent encoder-decoder with feedback. Figure~\ref{fig:variants} illustrates the 6 different network configurations that are evaluated here. 
We emphasize that our intention is not to put each encoder-decoder variant in competition, but to provide an illustrative comparison of the relative benefits of different configurations for the task being addressed here. Table~\ref{tab:discuss1} shows the performance of different variants on LSUN dataset.

The comparison of (a) and (b) variants indicates that stacking encoder-decoder networks can further improve the performance, as the network is enforced to learn the spatial structure of the room layout keypoints implicitly by placing constraints on multiple bottleneck layers.

However, adding skip connections~\cite{he2016deep, newell2016stacked} as in (c) does not improve the performance for this task. This could be because the size of the training set (thousands) is not as large as other datasets (millions) that have been evaluated on, therefore skipping layers is not necessary for the specific dataset.

Adding a feedback loop, implemented as a concatenation of input and previous prediction as a new input~\cite{tu2008auto, oberweger2015training} for the same encoder-decoder network as in (d) improves the performance. At each iteration, the network has access to the thus-far sub-optimal prediction along with the original input to help inference at the current time step.

Making an encoder-decoder recurrent with memory units (e) to behave as a RNN obtains the lowest keypoint error and pixel error (our full RoomNet model). The lateral connections in the recurrent encoder-decoder allow the network to carry information forward and help prediction at future time steps.
Finally, adding a feedback loop to the memory augmented recurrent encoder-decoder (f) does not improve the results. It is possible that using the memory augmented structure (e) can already store previous hidden state information well without feedback.
Note that weight matrices of the encoder-decoder are not shared in configurations (b) and (c) but shared in configurations (d), (e), and (f), resulting in more parametrically efficient architectures.

\begin{table}
\footnotesize{
\setlength{\tabcolsep}{3pt}
\def\arraystretch{1.2}
\center
\begin{tabular}{@{}l c c}
\toprule
\textbf{Model}  					& Keypoint Error (\%)  	&   Pixel Error (\%)  	\\
\midrule
Vanilla enc-dec (RoomNet basic)	&  6.95    			&  10.46			\\ 
Stacked enc-dec      				&  6.82      		&  10.31			\\ 
Stacked enc-dec with skip connect.	&  7.05      		&  10.48			\\  
Enc-dec w/ feedback				&  6.84    			&  10.10  			\\
Recurrent enc-dec (RoomNet full)	& \textbf{6.30}     	& \textbf{9.86}		\\
Recurrent enc-dec w/ feedback 	&  6.37    			&  9.88			\\
\bottomrule
\end{tabular}
\vspace{0.05cm}
\caption{Evaluation of encoder-decoder (enc-dec) variants on LSUN dataset~\cite{zhang2016large}. Note that recurrent encoder-decoders use 3 iteration time steps.}
\label{tab:discuss1}}
\vspace{-2mm}
\end{table}

\section{Conclusion}
\vspace{-1mm}
We presented a simple and direct formulation of room layout estimation as a keypoint localization problem. We showed that our RoomNet architecture and its extensions can be trained end-to-end to perform accurate and efficient room layout estimation. The proposed approach stands out from a large body of work using geometry inspired multi-step processing pipelines. In the future, we would like to adopt gating mechanism \cite{lee2016generalizing} to allow incoming signal to alter the state of recurrent units and extend RoomNet to sequential data for building room layout maps

{\small
\bibliographystyle{ieee}
\bibliography{egbib}
}

\end{document}